\documentclass[letterpaper, 10 pt, conference]{ieeeconf}  

\IEEEoverridecommandlockouts                              

\usepackage{graphicx,subfig}

\usepackage{hyperref}
\usepackage[sorting=none, style=ieee]{biblatex}
\usepackage{booktabs}  
\usepackage{threeparttable} 
\usepackage{makecell}
\usepackage{graphicx}
\usepackage{multicol}
\usepackage{multirow}
\usepackage{feyn}
\usepackage{amsmath,amssymb,amsfonts}
\usepackage{ulem}
\usepackage{caption}
\usepackage{fancyhdr} 
\usepackage{xcolor} 
\usepackage{float}
\usepackage{hyperref}

\hypersetup{
	colorlinks=true,
	linkcolor=cyan,
	filecolor=blue,      
	urlcolor=red,
	citecolor=green,
}

\title{\LARGE \bf
Annotation-Free Detection of Drivable Areas and Curbs Leveraging LiDAR Point Cloud Maps

}

\author{Fulong Ma, Daojie Peng, Jun Ma   
\thanks{Fulong Ma, Daojie Peng and Jun Ma are with The Hong Kong University of Science and Technology (Guangzhou), Guangzhou 511453, China (e-mail: \{fmaaf, dpeng108\}@connect.hkust-gz.edu.cn, jun.ma@ust.hk.)}
}

\begin{document}

\maketitle

\thispagestyle{empty}
\pagestyle{empty}

\begin{abstract}

Drivable areas and curbs are critical traffic elements for autonomous driving, forming essential components of the vehicle visual perception system and ensuring driving safety. Deep neural networks (DNNs) have significantly improved perception performance for drivable area and curb detection, but most DNN-based methods rely on large manually labeled datasets, which are costly, time-consuming, and expert-dependent, limiting their real-world application. Thus, we developed an automated training data generation module. Our previous work \cite{ma2023self} generated training labels using single-frame LiDAR and RGB data, suffering from occlusion and distant point cloud sparsity. In this paper, we propose a novel map-based automatic data labeler (MADL) module, combining LiDAR mapping/localization with curb detection to automatically generate training data for both tasks. MADL avoids occlusion and point cloud sparsity issues via LiDAR mapping, creating accurate large-scale datasets for DNN training. 
In addition, we construct a data review agent to filter the data generated by the MADL module, eliminating low-quality samples.
Experiments on the KITTI \cite{fritsch2013new}, KITTI-CARLA \cite{deschaud2021kitticarla} and 3D-Curb \cite{zhao2025curbnet} datasets show that MADL achieves impressive performance compared to manual labeling, and outperforms traditional and state-of-the-art self-supervised methods in robustness and accuracy.

\end{abstract}

\begin{figure*}[h]
    \setlength{\abovecaptionskip}{0pt}
    \setlength{\belowcaptionskip}{0pt}
    \centering
    \includegraphics[width=1.0\linewidth]{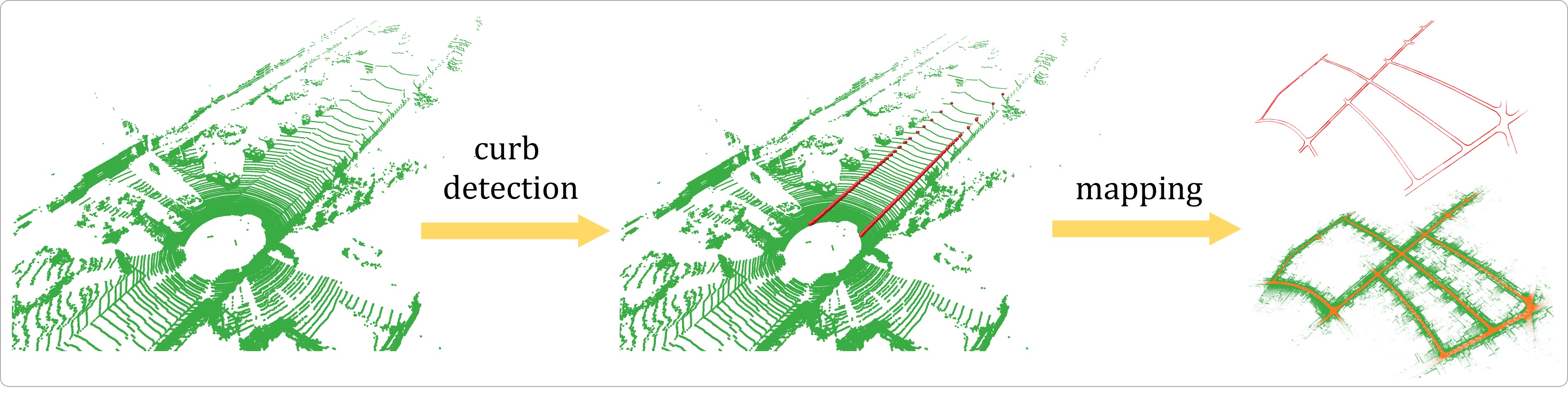}
    \captionsetup{font={small}}
    \caption{We use traditional methods to detect curbs from single-frame point cloud, and then leverage SLAM technology to construct curb maps and drivable area maps. These maps are used in subsequent steps to retrieve curb and drivable area information through localization within the map.}
    \label{mapping}
\end{figure*}

\section{INTRODUCTION}

Autonomous driving technology is fundamentally transforming human lifestyles, ranging from logistics and freight transportation to everyday mobility. However, to further advance autonomous driving systems to a higher level of maturity, numerous technical challenges still need to be addressed. The most fundamental question is how to achieve driving safety comparable to, or even exceeding, that of human drivers. Addressing this problem requires overcoming a series of critical technical barriers.
Among these challenges, driving scene understanding, particularly drivable area detection and curb detection, constitutes one of the core technical tasks in autonomous driving systems \cite{sless2019road,ma2025monocular}. In recent years, this field has achieved significant progress driven by breakthroughs in deep learning techniques \cite{alvarez2012road}. Although the mainstream direction of autonomous driving is increasingly shifting toward end-to-end paradigms \cite{chen2023e2esurvey}, as well as vision-language-action (VLA) models \cite{jiang2025survey} and world models \cite{tu2025drivingworldmodel}, scene understanding, including drivable area and curb detection, remains crucial for enhancing the interpretability and safety assurance of autonomous driving algorithms.

Current approaches to drivable area detection and curb detection can generally be categorized into two classes: traditional methods and learning-based methods. Traditional methods typically employ explicit geometric models to define drivable areas and curbs, and determine the optimal model parameters through optimization techniques \cite{fan2019pothole,wang2020speed}. For example, \cite{wedel2009b} proposed a representative traditional drivable area segmentation approach that performs road segmentation by fitting a B-spline model to the road disparity projection in a v-disparity image \cite{labayrade2002real}. In \cite{wang2020speed}, candidate points are extracted through multi-feature fusion, left and right boundaries are classified based on road segmentation lines, and curb detection is achieved using iterative Gaussian process regression.
In recent years, with the rapid advancement of deep learning, learning-based methods have achieved breakthrough improvements and are widely regarded as state-of-the-art solutions. For instance, Lu et al. \cite{lu2019monocular} adopted an encoder-decoder architecture to segment bird’s-eye-view RGB images, enabling end-to-end drivable area recognition. In \cite{zhao2025curbnet}, curb detection was addressed using deep learning techniques, achieving the best reported performance. Although learning-based techniques achieve excellent performance, annotating training data remains a very challenging problem.

To address the challenge of data annotation, some methods have explored annotation-free approaches. 
In paper \cite{mayr2018self}, drivable areas are separated from obstacles using the linear road features in the v-disparity map to create training data. Nevertheless, the method depends on stereo vision's limited accuracy and oversimplifies the drivable area as a single fitted line.
In paper \cite{ma2023self}, LiDAR point clouds are used to generate training labels by first detecting drivable areas within the point cloud and then projecting them onto the image plane. However, as this method relies on single-frame point clouds, it is susceptible to issues such as point sparsity, occlusions, and limited sensing range.

To boost annotation-free method performance, we introduce a novel module, the map-based automatic data labeler (MADL). It first applies traditional methods for ground segmentation and curb detection on single-frame point clouds. These detected points are then fused into a point cloud map via mapping, mitigating the sparsity, occlusion, and range limitations of single-frame data faced in \cite{ma2023self} and improving accuracy. Drivable areas and curb points are later retrieved from the map through point cloud localization and projected onto the image plane using camera parameters to generate training labels. Finally, a Data Review Agent is also designed to assess data quality and filter low-quality samples.


We evaluate our proposed approach using the KITTI dataset \cite{fritsch2013new}, the KITTI-CARLA dataset \cite{deschaud2021kitticarla}, and the 3D-Curb dataset introduced in \cite{zhao2025curbnet}. To validate the feasibility and effectiveness of the proposed MADL module, we assess the data generated by this module on these three datasets and train different deep neural network models using the training data produced by our MADL module. Experimental results demonstrate that the proposed MADL module achieves high-accuracy self-supervised drivable area and curb detection, and outperforms all existing methods.

Our main contributions are:
\begin{itemize}

\item We propose a novel automated training data generation method for drivable area and curb detection by combining single-frame point cloud processing with point cloud mapping.

\item We design a Data Review Agent to automatically evaluate the quality of the training data generated by MADL and filter out low-quality data.

\item We validate the effectiveness of the proposed MADL method on different datasets. Compared with manually annotated data, this method maintains competitive performance while significantly improving efficiency.

\end{itemize}

The rest of this paper is organized as follows: Section \ref{related_work} provides an overview of recent advances in the technologies relevant to this work. Section \ref{method} details the proposed method. Section \ref{experiments} presents the experimental results and discusses the effectiveness of the MADL module. Finally, Section \ref{conclusions} concludes the paper.

\begin{figure*}[h]
    \setlength{\abovecaptionskip}{0pt}
    \setlength{\belowcaptionskip}{0pt}
    \centering
    \includegraphics[width=1.0\linewidth]{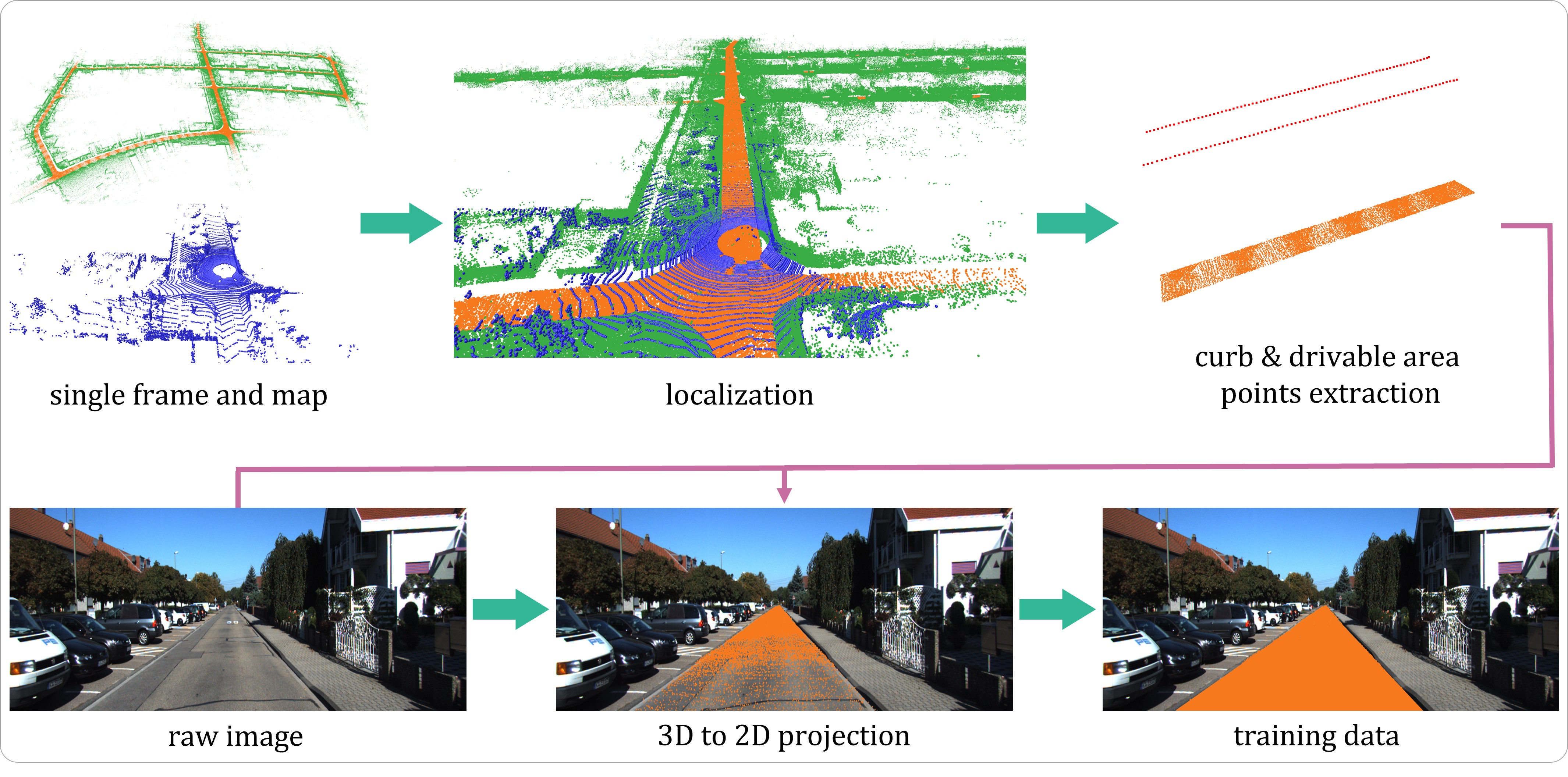}
    \captionsetup{font={small}}
    \caption{Given the point cloud input at a certain moment, the positional information at that moment is obtained through point cloud localization. Then, curb points and drivable area points within a specified range are retrieved from the map. The retrieved curb points can be directly used as training data for the curb detection task. The retrieved drivable area point cloud is projected onto the image plane using camera intrinsics, and through post-processing, training data for the drivable area detection task can be obtained.}
    \label{query}
\end{figure*}

\section{RELATED WORK}
\label{related_work}

\subsection{Drivable Area Detection}
For advanced driving assistant systems (ADAS) and autonomous driving vehicles, drivable area detection is a fundamental prerequisite for safe path planning and vehicle control. Traditional computer vision-based algorithms primarily rely on manually defined features, such as edge features \cite{yoo2013gradient} and histogram statistics \cite{son2015real}, which are tailored to specific scenarios and have poor adaptability to complex domains \cite{bar2014recent}. Yoo \textit{et al.} \cite{yoo2013gradient} proposed a gradient-enhancing method for illumination-robust segmentation, while Son \textit{et al.} \cite{son2015real} detected drivable areas via vanishing points, lane markers, and clustering. A pivotal development is the v-disparity map \cite{labayrade2002real}, which distinguishes ground planes and obstacles, with subsequent extensions optimizing its application \cite{soquet2007road,zhao2007global,yiruo2013complex}. With DNN advancements \cite{krizhevsky2017imagenet}, data-driven methods outperform traditional approaches, but most rely on massive, costly annotated data.

\subsection{Curb Detection}
Curb detection is a core perceptual task for autonomous driving, critical for defining drivable boundaries and ensuring path safety. With LiDAR’s popularization, its 3D point clouds have become the primary data source, with methods divided into traditional geometric feature-based and deep learning-based approaches. Traditional methods rely on manual features (e.g., height difference, smoothness): Zhang et al. \cite{zhang2018roadsegmentation} used road segmentation, Sun et al. \cite{sun20193d} optimized point cloud fitting, and Wang et al. \cite{wang2020speed} proposed a speed-accuracy framework, but all lack adaptability to complex scenes. Deep learning methods are mainstream: RangeNet++ \cite{milioto2019rangenet} enables fast segmentation, Cylinder3D \cite{zhou2020cylinder3d} enhances small-target detection, PVKD \cite{hou2022pointto} balances accuracy and real-time performance, and CurbNet \cite{zhao2025curbnet} achieves state-of-the-art results. 
However, large-scale, high-quality training data remains a significant challenge.

\subsection{Annotation-Free Method}

Automatic labeling can generate training data without human involvement, which is critical for reducing the training cost of DNNs. However, related research remains limited. Barnes \textit{et al.} \cite{barnes2017find} generated labels using odometry and obstacle sensing information, but their method suffers from odometry drift. Works \cite{mayr2018self, wang2019self} adopted the v-disparity map approach, where a line is fitted in the v-disparity map to represent drivable-area pixels. However, due to the inherent limitations of stereo cameras, the generated labels are of relatively low quality. In \cite{ma2023self}, Ma \textit{et al.} proposed using LiDAR point clouds to assist in generating training labels for drivable areas. Specifically, curb constraints are used to identify points belonging to drivable areas in the point cloud, which are then projected onto the image plane to generate training labels. This method significantly improves upon v-disparity map–based approaches. Nevertheless, it is still limited by the sparsity, occlusion, and restricted observation range of single-frame point clouds. Therefore, in this paper, we introduce mapping to address these limitations.

\subsection{Mapping and Localization}

Mapping and localization are key technologies for autonomous driving, providing spatial context and real-time pose estimation. Mapping aims to build an accurate environmental map for downstream localization, perception, and planning, while localization estimates vehicle state by minimizing the scan-matching distance between real-time LiDAR scans and a pre-built map.
LiDAR-based SLAM has advanced rapidly. For example, LOAM \cite{zhang2014loam} achieves high-precision real-time localization and mapping by separately estimating LiDAR odometry and map construction, leveraging edge and planar features, and lays the foundation for feature-based SLAM methods. 
LeGO-LOAM \cite{shan2018lego}, a ground-optimized variant of LOAM, efficiently and stably performs real-time LiDAR localization and mapping in complex environments.
The FAST-LIO series \cite{xu2021fast,xu2022fast} uses tightly-coupled optimization to fuse LiDAR and inertial measurement unit (IMU) data, achieving low-latency, high-precision, and robust real-time localization and mapping.
In this work, we do not focus on SLAM techniques themselves; rather, we leverage mapping and localization to better assist the generation of our training data.

\section{METHOD}
\label{method}

\subsection{LiDAR-Based Road Curb Detection}

We adopt the road curb detection algorithm introduced in Wang \textit{et al.} \cite{wang2020speed} to extract road information. This approach comprises four main stages: ground segmentation, feature extraction, feature classification, and feature filtering. A brief overview of each stage is given below.

\textit{(a) Ground Segmentation:} To optimize ground modeling, a piecewise plane fitting method is employed to distinguish ground points from non-ground points. The raw point cloud frame $P$ is divided into multiple segments along the x-axis, and a plane fitting algorithm \cite{zermas2017fast} is applied to each segment to extract ground points $P_g$ and non-ground points $P_{ng}$.

\textit{(b) Feature Extraction:} Three spatial features are used to extract feature points. Ground points $P_g$ are divided into scan layers equal to the number of sensor rings, with each layer containing points from the corresponding laser. Let $P_{ri} = [x_{ri},y_{ri},z_{ri}]$ represent a point, where $r$ is the ring ID. The three spatial features are described as follows:

\begin{itemize}
    \item {Height Difference:}
    Let $Z_{max}$ and $Z_{min}$ denote the maximum and minimum $z$-values among the neighbors of $P_{ri}$, respectively. The height difference feature is defined as:
    \begin{equation}
        H_{1} \leq Z_{max} - Z_{min} \leq H_{2},
    \end{equation}
    \begin{equation}
        \sqrt{\frac{\sum(z_{ri} - \mu)^{2}}{n_{neighbor}}} \geq H_{3},
    \end{equation}
    where $n_{neighbor}$ is the number of neighbors, $z_{ri}$ is the $z$ value of each neighbor, $H_{1}$, $H_{2}$, and $H_{3}$ are thresholds. Also, $\mu =\sum z_{ri} / n_{neighbor}$.
   
    \item {Smoothness:}
     This feature is defined as:
     \begin{equation}
         s = \frac{\big \| \sum_{P_{ri} \in S, j \neq i} (p_{ri} - p_{rj})  \big \|}{|S|\cdot||P_{ri}||},
     \end{equation}
     \begin{equation}
         s \geq T_{s},
     \end{equation}
     where $s$ is the smoothness value of $p_{ri}$, $|S|$ is the cardinality of $S$, and $T_{s}$ is the threshold.
    \item {Horizontal Distance:}
    This feature is defined as:
    \begin{equation}
        \delta_{xy, r} = H_{s} \cdot \cot \theta_{r} \cdot \pi\theta_{a},
    \end{equation}
    where $H_{s}$ is the absolute value of the height of point $p_{ri}$, $\theta_{l}$ is the vertical azimuth of scanning layer $r$, $\theta_{a}$ is the horizontal angular resolution of LiDAR. Note that $\theta_{r}$ and $\theta_{a}$ are both in radians.
\end{itemize}

\textit{(c) Feature Classification:}
Feature points are classified using a road-segmentation-line-based method. A beam model is built from non-ground points $P_{ng}$, and the longest beams in the front and rear regions define the road direction. After outlier removal via peak detection, feature points are separated into left and right groups based on the segmentation line.

\textit{(d) Feature Filtering:}
After feature extraction and classification, some false points remain. To remove them, an iterative Gaussian Process Regression algorithm is applied to model road boundaries and eliminate outliers.

Due to the sparsity and occlusion of single-frame point clouds at far ranges, we apply spatial truncation before curb detection. Specifically, we crop the point cloud along the LiDAR x-axis to retain only the high-density region within 30 meters in front of sensor.

\begin{figure*}[h]
    \setlength{\abovecaptionskip}{0pt}
    \setlength{\belowcaptionskip}{0pt}
    \centering
    \includegraphics[width=1.0\linewidth]{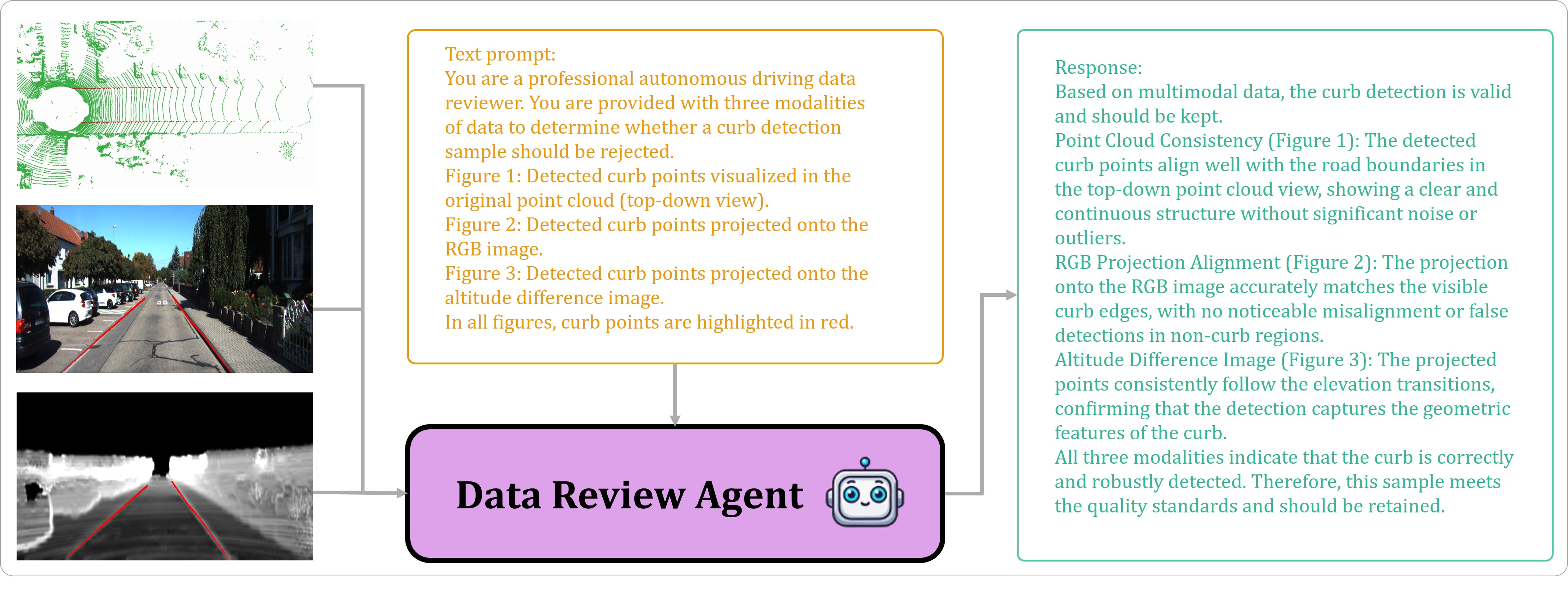}
    \captionsetup{font={small}}
    \caption{The schematic diagram of our Data Review Agent. It takes as input three images from different modalities (a point cloud top-down view, an RGB image, and an altitude difference image), each containing curb detection information. Based on the quality of the curb detection, the agent then outputs whether the detection data from that frame should be retained. }
    \label{DRA}
\end{figure*}

\subsection{Construct Curb and Drivable Area Map}

As discussed above, single-frame point clouds are inherently limited by sparse distant points, susceptibility to occlusion, and restricted sensing range. To mitigate these issues and improve curb detection accuracy, we apply truncation to the point clouds during the curb detection stage. Furthermore, to acquire drivable area and curb points beyond the observable range of a single frame, as illustrated in Fig. \ref{mapping}, we aggregate the curb and drivable area points obtained from the truncated point cloud processing into a global map $\mathcal{M}$ using SLAM techniques \cite{xu2021fast,zhang2014loam}. Once the pose of a specific frame is known, more accurate and longer-range drivable area and curb points can be retrieved from the map.

\subsection{Extract Drivable Area and Curb Points from the Map}
In the previous stage, road curb and drivable area information was incorporated into the map via SLAM. Extracting these semantic features requires accurate localization of the current frame within map $\mathcal{M}$. Specifically, point cloud registration is utilized to align the real-time scan $\mathcal{P}_s = \{p_i\}_{i=1}^n$ with the prior map $\mathcal{M}$, a process that manifests as a non-linear least-squares optimization on the Special Euclidean Group $SE(3)$. The goal is to find the optimal transformation $\mathbf{T} = \{\mathbf{R}, \mathbf{t}\}$ that minimizes the geometric residuals between the transformed points and their map correspondences $q_i \in \mathcal{M}$:

\begin{equation}
\mathbf{T}^* = \arg \min_{\mathbf{T} \in SE(3)} \sum_{i=1}^{n} \omega_i \left\| (\mathbf{R} p_i + \mathbf{t}) - q_i \right\|^2
\end{equation}

where $\mathbf{R}$, $\mathbf{t}$, and $\omega_i$ represent the rotation matrix, translation vector, and weight factor, respectively. Iterative optimization ensures that cumulative motion drift is eliminated, achieving centimeter-level precision. Based on this robust localization, the system can then precisely query curb and drivable area points from $\mathcal{M}$, as illustrated in Fig. \ref{query}.

\subsection{Project 3D Drivable Area Points to Image Plane}
\label{D}

We employ the pinhole camera model to project 3D points onto the 2D image plane. This process involves two main steps: first, the point cloud coordinates are transformed into the image’s world coordinate system; second, these 3D points are projected onto the image plane using the camera intrinsic parameters, yielding the corresponding 2D points that lie within the drivable area.



The projection of a 3D point $ \textbf{x} = (x, y, z, 1)^{T} $ in rectified (rotated) camera coordinates to a point $ \textbf{y} = (u, v, 1)^{T} $ in the camera image plane is given as: 

\begin{equation} \label{eq1}
    \textbf{y} = \textbf{P}_{rect} \cdot \textbf{x}
\end{equation}
with

\begin{equation}  \label{eq2}
    \textbf{P}_{rect} =
	\begin{pmatrix} f_{u}  & 0     & c_{u} & -f_{u}b_{x} 
                      \\ 0 & f_{v} & c_{v} & 0 
                      \\ 0 & 0     & 1     & 0     
    \end{pmatrix}
\end{equation}

the projection matrix. Here, $f_{u}$, $f_{v}$ are focal length in pixels, $c_{u}$, $c_{v}$ are optical center in pixels, $b_{x}$ denotes the baseline with respect to reference camera.


The rigid body transformation from LiDAR's coordinate to camera's coordinate is given by rotation matrix $\textbf{R}_{LiDAR}^{camera} \in \mathbb{R}_{3\times 3}$ and translation vector $ \textbf{t}_{LiDAR}^{camera} \in \mathbb {R}^{1 \times 3} .$ 

Using

\begin{equation} \label{eq4}
    \textbf{T}_{LiDAR}^{camera} =
	\begin{pmatrix}    \textbf{R}_{LiDAR}^{camera}  & \textbf{t}_{LiDAR}^{camera}      
                \\                              0   & 1     
    \end{pmatrix}
\end{equation}

a 3D point $\textbf{x}$ in LiDAR's coordinate gets projected to a point $\textbf{y}$ in the camera image plane as:
\begin{equation} \label{eq5}
    \textbf{y} = \textbf{P}_{rect} \cdot  \textbf{T}_{LiDAR}^{camera} \cdot \textbf{x}.
\end{equation}

In this way, we project the drivable area points in the LiDAR coordinate system to the image plane, and then we obtain the training data labels by finding the concave hull for these projected points. 
\textcolor{black}{It should be noted that for curb training data generation, since our evaluation is conducted using point cloud-based models directly on the point cloud, we only extract the points belonging to the curb from the map $\mathcal{M}$, without further projecting them onto the image plane.}

\begin{table}[h]
\setlength{\tabcolsep}{3.5pt}
\captionsetup{font={footnotesize}}
\caption{
The comparison results of our MADL with TDG \cite{mayr2018self} / SSLG \cite{wang2019self} and ADL \cite{ma2023self} on the KITTI dataset.}
\label{kitti_1}
\begin{center}

\begin{tabular}{c c c c c c} 
\hline
Method           & Accuracy $\uparrow$ &Precision $\uparrow$ &Recall $\uparrow$ & F1-score $\uparrow$& IoU $\uparrow$ \\
\hline
 TDG  / SSLG        &87.81  & 59.59 & 84.03 &  69.73  &  53.54 \\

 ADL    & 93.12  & 90.68 & 91.78 & 91.23 & 85.36 \\
 MADL (Ours)  & \textbf{97.34}   &\textbf{93.92}  & \textbf{94.78} & \textbf{94.35} & \textbf{92.29} \\
\hline
\end{tabular}
\end{center}
\end{table}

\begin{table}[h]
\setlength{\tabcolsep}{3.5pt}
\captionsetup{font={footnotesize}}
\caption{The comparison results of our MADL with TDG \cite{mayr2018self} / SSLG \cite{wang2019self} and ADL \cite{ma2023self} on the KITTI-CARLA dataset.}
\label{kitti_carla_1}
\begin{center}
\begin{tabular}{c c c c c c} 
\hline
Method           & Accuracy $\uparrow$ &Precision $\uparrow$ &Recall $\uparrow$ & F1-score $\uparrow$& IoU $\uparrow$ \\
\hline
 TDG  / SSLG    &88.79  & 79.52 & 82.41 &  80.94  &  68.01 \\

 ADL    & 94.13  & 91.31 & 92.38 & 91.84 & 86.24 \\
 MADL (Ours)  & \textbf{97.25}   &\textbf{94.42}  & \textbf{95.84} & \textbf{95.13} & \textbf{93.87} \\
\hline
\end{tabular}
\end{center}
\end{table}

\begin{table*}[t]
\renewcommand\arraystretch{1.0}
\captionsetup{font={footnotesize}}
\captionsetup{
  width=1.0\textwidth, 
}
\caption{Comparison of training results of three semantic segmentation models—U-Net \cite{ronneberger2015u}, SegFormer \cite{xie2021segformer}, and VM-UNet \cite{ruan2024vm}—on the training data generated by TDG \cite{mayr2018self} / SSLG \cite{wang2019self}, ADL \cite{ma2023self}, and our MADL module on the KITTI dataset, as well as on the ground truth. The results trained on ground truth are marked in teal and are not compared with the baseline methods, they are provided for reference only.}
\label{kitti_2}
\begin{center}
\setlength{\tabcolsep}{4.0mm}{
\begin{tabular}{c c c c c c c c} 
\toprule
Network  & Network Architecture    & Method     & Accuracy $\uparrow$ &Precision $\uparrow$ &Recall $\uparrow$ & F1-score $\uparrow$& mIoU $\uparrow$ \\

\hline
\multirow{4}{*}{U-Net } &\multirow{4}{*}{CNN} 

& TDG  / SSLG  &90.13 &80.15 &\textbf{83.32} &82.18  &71.80 \\
 & & ADL      &91.43  &86.12 &81.65 &83.83 &73.96 \\
 &  & MADL (Ours)     &\textbf{92.76} &\textbf{87.93} &82.03 &\textbf{84.88} &\textbf{74.02} \\
  & & \textcolor{teal}{Ground Truth}     &\textcolor{teal}{95.56} &\textcolor{teal}{89.86} &\textcolor{teal}{82.95} &\textcolor{teal}{86.27} &\textcolor{teal}{75.13} \\
\hline
\multirow{4}{*}{SegFormer} &\multirow{4}{*}{Transformer} 
& TDG  / SSLG &90.43 &81.43  & 84.02 & 82.70 & 72.11 \\
 & &ADL     &92.44 &88.84  & 90.29 &89.56  &85.71  \\
 &  & MADL (Ours) & \textbf{96.35} &\textbf{92.76} &\textbf{93.13} &\textbf{92.95} & \textbf{89.91} \\
   & & \textcolor{teal}{Ground Truth}     & \textcolor{teal}{98.05}  &\textcolor{teal}{95.22} &\textcolor{teal}{95.65} & \textcolor{teal}{95.43} &\textcolor{teal}{91.68} \\
\hline
\multirow{4}{*}{VM-UNet } &\multirow{4}{*}{Mamba} 

& TDG  / SSLG &89.87 &81.12 &82.26  &81.69 &71.67\\
 & & ADL    &92.35  & 87.72 & 90.07 & 88.88 &85.55  \\

 & & MADL (Ours) & \textbf{95.93} &\textbf{92.74} &\textbf{91.67} &\textbf{92.20} & \textbf{87.58} \\
   & & \textcolor{teal}{Ground Truth}     & \textcolor{teal}{97.34} &\textcolor{teal}{94.45} &\textcolor{teal}{94.63} & \textcolor{teal}{94.54} &\textcolor{teal}{89.77} \\

\toprule
\label{single_modal}
\end{tabular}}
\end{center}
\end{table*}

\subsection{Data Review Agent}
Although point cloud truncation is applied to preserve sufficient point density within a single frame, and mapping is introduced to compensate for the limited spatial coverage of individual scans, curb detection can still fail in challenging scenarios, such as severe occlusion or weak geometric cues. These failure cases inevitably introduce noisy or invalid samples into the automatically generated dataset, which should be excluded to ensure training reliability.
To this end, we develop a Data Review Agent for automatic data filtering. The agent is built upon large foundation models \cite{ma2026safety,bai2025qwen3}, leveraging their strong generalization and multimodal reasoning capabilities. For each sample, we construct three complementary modalities: an RGB image, an Altitude Difference Image (ADI), and a top-down (bird’s-eye-view) visualization of the point cloud. The curb points obtained from the global map are projected onto these modalities. For the RGB image and ADI, projection is performed following the method described in Section \ref{D}, while for the top-down point cloud view, the mapped curb points are directly overlaid onto the bird’s-eye visualization.
The resulting multimodal representations, augmented with projected curb annotations, are then fed into the Data Review Agent, which determines whether the sample should be retained or discarded, as illustrated in Fig. \ref{DRA}.

\begin{figure}[h]
    \setlength{\abovecaptionskip}{0pt}
    \setlength{\belowcaptionskip}{0pt}
    \centering
    \includegraphics[width=1.0\linewidth]{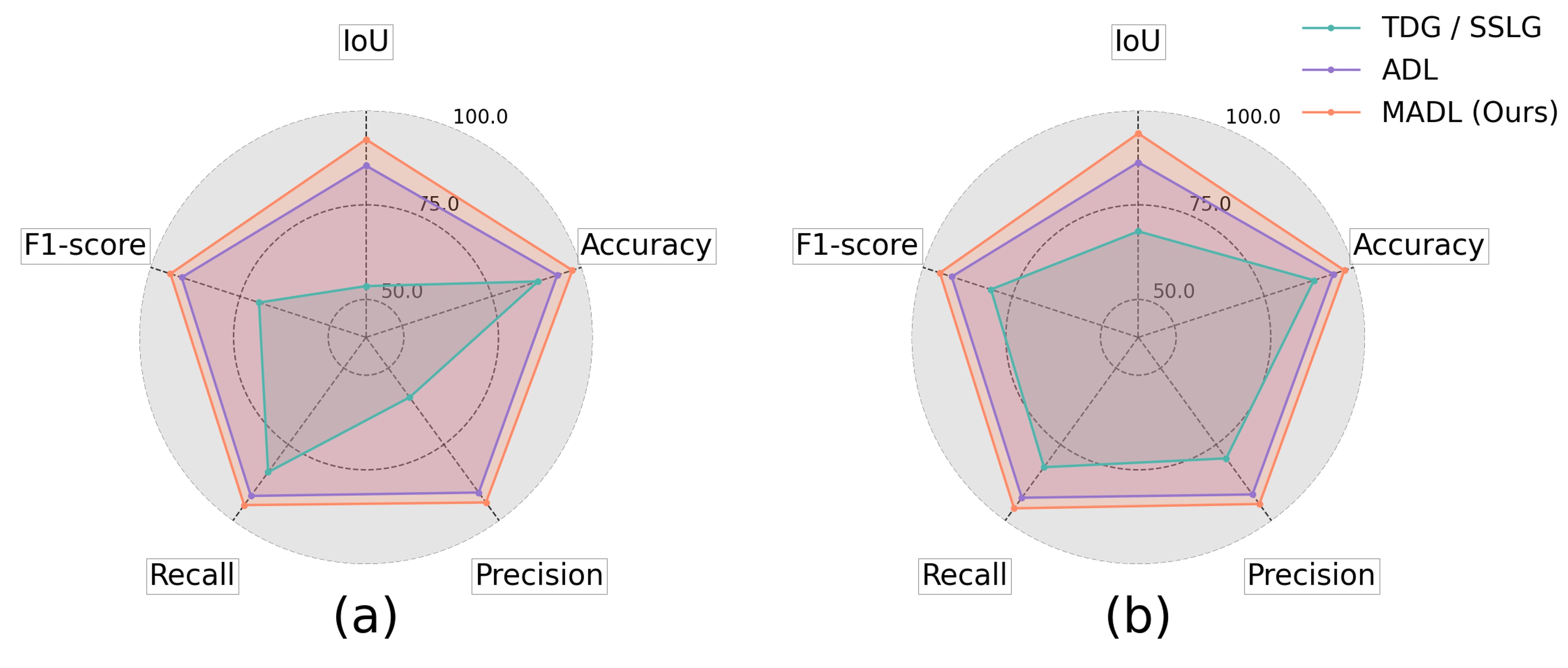}
    \captionsetup{font={small}}
    \caption{The qualitative comparison results of our MADL with TDG \cite{mayr2018self} / SSLG \cite{wang2019self} and ADL \cite{ma2023self} on the KITTI and KITTI-CARLA datasets. (a): Comparison on the KITTI dataset. (b): Comparison on the KITTI-CARLA dataset.}
    \label{radar_plot}
\end{figure}

\begin{table*}[h]
\renewcommand\arraystretch{1.0}
\captionsetup{font={footnotesize}}
\captionsetup{
  width=1.0\textwidth, 
}
\caption{Comparison of training results of three semantic segmentation models—U-Net \cite{ronneberger2015u}, SegFormer \cite{xie2021segformer}, and VM-UNet \cite{ruan2024vm}—on the training data generated by TDG \cite{mayr2018self} / SSLG \cite{wang2019self}, ADL \cite{ma2023self}, and our MADL module on the KITTI-CARLA dataset, as well as on the ground truth. The results trained on ground truth are marked in teal and are not compared with the baseline methods, they are provided for reference only.}
\label{table_1}
\begin{center}
\setlength{\tabcolsep}{4.0mm}{
\begin{tabular}{c c c c c c c c} 
\toprule
Network  & Network Architecture    & Method     & Accuracy $\uparrow$ &Precision $\uparrow$ &Recall $\uparrow$ & F1-score $\uparrow$& mIoU $\uparrow$ \\

\hline
\multirow{4}{*}{U-Net } &\multirow{4}{*}{CNN} 

& TDG  / SSLG  &89.36 &80.68 &83.97 &82.29  &72.82 \\
 & & ADL      &91.32  &87.75 &88.36 &88.05 &78.53 \\
 &  & MADL (Ours)     &\textbf{92.33} &\textbf{89.12} &\textbf{90.89} &\textbf{90.00} &\textbf{83.44} \\
  & & \textcolor{teal}{Ground Truth}   &\textcolor{teal}{93.03}  &\textcolor{teal}{90.22} &\textcolor{teal}{92.75} &\textcolor{teal}{91.47} &\textcolor{teal}{84.12} \\
\hline
\multirow{4}{*}{SegFormer } &\multirow{4}{*}{Transformer} 
& TDG  / SSLG  &90.56 &81.32 &84.37 &82.82 & 73.58\\
 & &ADL      & 93.65 & 89.17 & 90.43 & 89.80 & 86.12 \\
 &  & MADL (Ours) & \textbf{95.71} &\textbf{92.98} &\textbf{93.23} &\textbf{93.10}  & \textbf{90.78} \\
 & & \textcolor{teal}{Ground Truth}   & \textcolor{teal}{97.98} &\textcolor{teal}{94.56} &\textcolor{teal}{95.43}&\textcolor{teal}{94.99} &\textcolor{teal}{92.31} \\
\hline
\multirow{4}{*}{VM-UNet} &\multirow{4}{*}{Mamba} 

& TDG  / SSLG  &89.78 &81.12  &84.45  &82.75  & 73.46\\
 & & ADL    &93.17  & 89.78 & 90.45 & 90.11  & 85.37 \\

 & & MADL (Ours) & \textbf{95.65} &\textbf{92.78} &\textbf{93.33} &\textbf{93.05} & \textbf{90.06} \\
 & & \textcolor{teal}{Ground Truth}   & \textcolor{teal}{97.12} & \textcolor{teal}{94.43} &\textcolor{teal}{94.78}  &  \textcolor{teal}{94.60}&\textcolor{teal}{91.45}  \\

\toprule
\label{kitti_carla_2}
\end{tabular}}
\end{center}
\end{table*}

\section{EXPERIMENT}
\label{experiments}

\subsection{Datasets and Experimental Setup}

In our experiments, we evaluate the performance of our proposed method on three datasets. For the drivable area detection task, we conduct experiments using the KITTI road dataset \cite{fritsch2013new} and the KITTI-CARLA dataset \cite{deschaud2021kitticarla}. For the curb detection task, we employ the 3D-Curb dataset \cite{zhao2025curbnet}.

For the drivable area detection task, we first compare the training data generated by our MADL module against the ground truth of the KITTI road dataset and the KITTI-CARLA dataset, respectively. Subsequently, we utilize the training data generated by our MADL, two baseline methods (TDG / SSLG \cite{mayr2018self,wang2019self} and ADL \cite{ma2023self}), and the official ground truth to train three segmentation models with distinct architectures: U-Net \cite{ronneberger2015u}, SegFormer \cite{xie2021segformer}, and VM-UNet \cite{ruan2024vm}. The training performance of these three models is then compared.

For the curb detection task, we train three point cloud segmentation models—PVKD \cite{hou2022point}, Cylinder3D \cite{zhou2020cylinder3d}, and CurbNet \cite{zhao2025curbnet}—using the curb training data generated by our MADL and the 3D-Curb dataset. We then evaluate and compare the training performance of these three point cloud segmentation models on both the training data generated by our MADL and the original 3D-Curb dataset, using the labels from the 3D-Curb dataset as the evaluation benchmark.

\subsection{Evaluation Metrics}

For drivable area detection task, we adopt five widely used evaluation criteria: Accuracy, Precision, Recall, F1-score, and Intersection over Union (IoU). These metrics capture overall prediction correctness, positive prediction reliability, detection completeness, and region overlap quality, respectively. They are defined as:
$
Accuracy =   \frac{N_{TP} + N_{TN}}{N_{TP} + N_{FP}  + N_{TN}  + N_{FN}}, 
$
$
Precision =  \frac{N_{TP}}{N_{TP}+N_{FP}},
$
$
Recall =  \frac{N_{TP}}{N_{TP} + N_{FN}},
$
$
F1-score =  \frac{2 * Precision * Recall}{Precision + Recall},
$ 
$
IoU =  \frac{N_{TP}}{ N_{TP} + N_{FP} + N_{FN}}.
$
Here, $N_{TP}$, $N_{TN} $, $N_{FP}$ and $N_{FN}$ denote the numbers of true positive, true negative, false positive, and false negative pixels, respectively.
For curb detection, we follow the metrics used in the previous work CurbNet \cite{zhao2025curbnet}, namely Precision, Recall, and F1-Score.

\subsection{Performance Evaluation}

In this section, we discuss the experimental results of the two tasks on three datasets. For the drivable area detection task, we conducted extensive experiments on the KITTI, KITTI-CARLA datasets.

We first compared the training data generated by our MADL and the training data generated by the two baseline methods, TDG/SSLG and ADL, against the ground truth of the KITTI and KITTI-CARLA datasets. The qualitative results are shown in Fig. \ref{radar_plot}, and the quantitative experimental results are presented in Tables \ref{kitti_1} and \ref{kitti_carla_1}. From the figure and the two tables, it can be observed that our MADL comprehensively surpasses the existing annotation-free methods TDG/SSLG and ADL across all five metrics, with a significant lead. This demonstrates the high quality of the training data generated by our MADL for drivable area detection, which is much closer to the ground truth.
Next, we trained semantic segmentation models with three different architectures (CNN, Transformer, and Mamba) using the training data generated by our MADL. We also trained these three models using the training data generated by TDG/SSLG and ADL, and then compared their performance on the test set. The experimental results are shown in Tables \ref{kitti_2} and \ref{kitti_carla_2}, with Table \ref{kitti_2} presenting the experiments on the KITTI dataset and Table \ref{kitti_carla_2} on the KITTI-CARLA dataset. Additionally, we report the performance of these three models trained on the official ground truth training set, marked in teal in the tables.  
From Tables \ref{kitti_2} and \ref{kitti_carla_2}, it can be seen that the three semantic segmentation algorithms trained on the dataset generated by our MADL module achieved the best performance, leading across all metrics compared to TDG/SSLG and ADL. When compared to the results trained on ground truth, our method also achieves comparable performance.  
Overall, Tables \ref{kitti_1}, \ref{kitti_carla_1}, \ref{kitti_2}, and \ref{kitti_carla_2} demonstrate the effectiveness and superiority of our MADL, as well as the high quality of the drivable area detection training data generated by the MADL module.

Considering both the drivable area detection task and the curb detection task, it can be observed that our proposed MADL demonstrates effectiveness and superiority in automatically generating data for these two tasks. Without requiring any manual annotation, our method can generate high-quality training data at scale for drivable area detection and curb detection tasks, which is of great significance for practical applications in autonomous driving.

\begin{table}[h]
\renewcommand\arraystretch{1.0}
\centering
\caption{
Performance comparison of four learning-based models—PVKD \cite{hou2022point}, Cylinder3D \cite{zhou2020cylinder3d}, and CurbNet \cite{zhao2025curbnet}—after training on the data generated by our MADL versus the 3D-Curb dataset \cite{zhao2025curbnet}, with a tolerance error of 0.2 meters.
}
\label{tab:curb_detect}
\setlength{\tabcolsep}{5.5pt}
\begin{tabular}{c|c|ccc}
\toprule
Method & Training Data 
& Precision $\uparrow$ 
& Recall $\uparrow$ 
& F1-Score $\uparrow$ \\
\midrule
\multirow{2}{*}{PVKD }
& 3D-Curb & \textbf{93.98} & 94.17 & \textbf{94.07}  \\
& MADL (Ours) &93.35  & \textbf{94.53} & 93.94 \\
\midrule
\multirow{2}{*}{CurbNet }
& 3D-Curb & \textbf{97.38} &95.01  & \textbf{96.17} \\
& MADL (Ours) &96.87  & \textbf{95.31} & 96.08 \\
\midrule
\multirow{2}{*}{Cylinder3D }
& 3D-Curb & \textbf{93.60} & \textbf{93.55} &\textbf{92.57} \\
& MADL (Ours) & 91.98 & 92.39 & 92.18 \\
\bottomrule
\end{tabular}
\end{table}

For the curb detection task, we trained three learning-based methods—PVKD, Cylinder3D, and CurbNet—on both the training data generated by our MADL and the 3D-Curb dataset, using the labels from the 3D-Curb data as ground truth, and compared their performance. The results are shown in Table \ref{tab:curb_detect}. 
It can be seen that the three models trained on our MADL-generated curb detection data achieved slightly lower results, with a performance decrease of less than 0.5\% in the F1-Score metric. Moreover, in terms of the Recall metric, the PVKD and CurbNet models trained on our data achieved leading performance., demonstrating the high quality of the curb detection training data generated by our MADL. It is worth mentioning that the annotation of the 3D-Curb dataset relies on human annotators, whereas our MADL requires zero human involvement.

\begin{table}[h]
\centering
\caption{Ablation study on the effectiveness of the Data Review Agent conducted on the KITTI dataset.}
\label{tab:ablation_study}
\footnotesize
\setlength{\tabcolsep}{2pt}
\begin{tabular}{c|ccccc}
\toprule
 Data Review Agent 
& Accuracy $\uparrow$ 
& Precision $\uparrow$ 
& Recall $\uparrow$ 
& F1-score $\uparrow$ 
& IoU $\uparrow$ \\
\midrule
 w/o &95.63  &92.17  &93.15  & 92.66 & 90.38 \\
 w/ &\textbf{97.34}  &\textbf{93.92}  &\textbf{94.78} &\textbf{94.35}  &\textbf{92.29} \\
\bottomrule
\end{tabular}
\end{table}

\subsection{Ablation Study}
To validate the effectiveness of our Data Review Agent, we conducted an ablation study on the KITTI dataset. The experimental results, as shown in Table \ref{tab:ablation_study}, demonstrate that with the Data Review Agent employed, our final training data achieves leading performance across all metrics compared to the ground truth, which fully illustrates the effectiveness of our Data Review Agent.

\section{CONCLUSIONS}
\label{conclusions}
This paper proposes MADL, a novel module for automatically generating large-scale training datasets for drivable area detection and curb detection. By truncating single-frame point clouds and performing curb detection within these truncated high-density point cloud regions, the adverse effects of point cloud sparsity are effectively mitigated. Furthermore, to extend beyond the observation range of a single-frame point cloud, we employ SLAM technology to integrate the drivable area information and curb information extracted from single-frame point clouds into a global map. Subsequently, through point cloud localization, the positional information of each single-frame datum is obtained, enabling the retrieval of drivable area points and curb points from the global map. High-quality training labels for drivable area detection and curb detection are then generated through post-processing. In addition, we introduce a Data Review Agent to filter out low-quality data samples produced by MADL, further enhancing the quality of the generated training data. Through the method proposed in this paper, large-scale training data for drivable area detection and curb detection can be obtained without the involvement of human annotators, providing a practical and low-cost solution for autonomous driving data engineering.



\defbibenvironment{bibliography}
  {\list
     {\printtext[labelnumberwidth]{%
        \printfield{labelprefix}%
        \printfield{labelnumber}}}
     {\setlength{\labelwidth}{\labelnumberwidth}%
      \setlength{\leftmargin}{\labelwidth}
      \addtolength{\leftmargin}{\labelsep}
      \setlength{\itemsep}{\bibitemsep}
      \setlength{\parsep}{\bibparsep}}%
      \renewcommand*{\makelabel}[1]{##1\hss}} 
  {\endlist}
  {\item}

\normalem
\renewcommand{\bibfont}{\small}
\setlength{\labelsep}{0.25em}  
\printbibliography

\end{document}